\documentclass{article}

    \usepackage[final]{neurips_2023}

\usepackage[utf8]{inputenc} 
\usepackage[T1]{fontenc}    
\usepackage{hyperref}       
\usepackage{url}            
\usepackage{booktabs}       
\usepackage{amsfonts}       
\usepackage{nicefrac}       
\usepackage{microtype}      
\usepackage{xcolor}         
\usepackage{graphicx}
\usepackage{multirow}
\usepackage{listings} 
\usepackage{xcolor} 
\usepackage{pifont}
\usepackage{authblk}

\newcommand{\cmark}{\ding{51}}%
\newcommand{\xmark}{\ding{55}}%

\definecolor{cadmiumgreen}{rgb}{0.0, 0.42, 0.24}
\definecolor{celestialblue}{rgb}{0.29, 0.59, 0.82}

\newcommand{\myblue}[1]{\textcolor{celestialblue}{#1}}
\title{OverPrompt: Enhancing ChatGPT through Efficient In-Context Learning}

\author[*1]{Jiazheng Li}
\author[*1]{Runcong Zhao}
\author[3]{Yongxin Yang}
\author[1,2]{Yulan He}
\author[1]{Lin Gui}
\affil[1]{Department of Informatics, King's College London, UK}
\affil[2]{The Alan Turing Institute, UK} 
\affil[3]{Queen Mary University of London, UK}

\begin{document}
\maketitle
\def\thefootnote{*}\footnotetext{Equal contribution. Email: \{jiazheng.li, runcong.zhao\}@kcl.ac.uk}\def\thefootnote{\arabic{footnote}}

\begin{abstract}

The remarkable performance of pre-trained large language models has revolutionised various natural language processing applications. Due to huge parameter sizes and extensive running costs, companies or organisations tend to transfer the models to the target task by zero-shot prompting techniques.   
However, the prohibitive costs of tokens and time have hindered their adoption in applications. 
We propose OverPrompt, leveraging the in-context learning capability of LLMs to handle multiple task inputs, thereby reducing token and time costs. This approach could potentially improve task performance during API queries due to better conditional distribution mapping. 
Evaluated across diverse classification datasets, our experiments show that OverPrompt can achieve cost-efficient zero-shot classification without causing significant detriment to task performance, and in some cases, even improving it. 
An ablation study conducted on various LLMs, along with an investigation into the robustness of our prompting strategy to different input ordering, offers valuable insights into the broader applicability of our method across diverse tasks. These findings also suggest a more seamless integration of our method with LLMs through an API.
\end{abstract}

\section{Introduction}
Large Language Models (LLMs), such as ChatGPT \citep{Ouyang2022TrainingLM}, have shown emergent abilities on various Natural Language Processing (NLP) tasks \citep{Wei2022EmergentAO}. In-context learning (ICL) approaches, including zero-shot or few-shot prompting strategies \citep{kojima2022large, lm_few_shot_learners, scalinglaws}, offering computational efficiency and considerable performance gains without the need of fine-tuning LLMs. 
Various research has been carried out to improve the performance of zero/few-shot learning, such as searching for better few-shot examples \citep{zhang2023automatic,Wang2023LargeLM} or finding appropriate prompts \citep{wang2023large, white2023prompt}.

However, cost-efficient prompting strategies are under-explored. Given the immense parameter size of LLMs, 
deploying them locally for certain industrial applications becomes impractical. Consequently, the substantial time and token costs associated with accessing these models through APIs present a significant challenge when adopting these models in production environments.  
While few-shot prompting entails the inclusion of demonstration examples, thereby raising the token cost of API queries, in most zero-shot prompting cases \citep{mialon2023augmented, white2023prompt}, the bulk of the input content is allocated to the task description, leaving only a portion for the task input. The repetition of the task description can result in a substantial cumulative cost for each individual query. 
Hence, it becomes imperative to reduce the token and time costs associated with utilising these LLMs. 


To address this issue, we propose OverPrompt, a zero-shot prompting strategy designed to process multiple instances simultaneously in a single query to enhance efficiency. Leveraging the emergent capability of LLMs, known as ICL, we analyse our prompting strategy within a Bayesian inference framework. 
Theoretically, our prompting strategy ensures better approximation of input task distributions by incorporating additional data and mitigating format errors. We also empirically show that our designed sampling and formatting framework enhances performance.
In order to understand the overall impact of OverPrompt on query efficiency, we evaluate OverPrompt across ten different text classification datasets. Our experiments reveal that OverPrompt reduces both token and time costs, while leveraging the in-context learning capabilities of LLMs to produce improved conditional distributions for tasks when additional instances are provided. 
We also modify the output formatting to address performance degradation and reduce errors. 
Performance enhancements are observed when contextual information supplements the model's decision-making process. This is particularly useful in tasks such as fact-checking, where extra evidence or logical deductions can be provided, and sentiment analysis, where well-defined category boundaries can be established through comparison. Nevertheless, tasks like sentence entailment may not gain any advantage from such context input\footnote{Our code is avaliable at \url{https://github.com/lijiazheng99/OverPrompt}.}.

\section{OverPrompt}
In this section, we introduce OverPrompt, a zero-shot classification strategy that utilizes ChatGPT's ICL ability for more efficient zero-shot classification, reducing token consumption and time costs.

\begin{figure*}[htb]
    \centering
    \includegraphics[width=0.90\linewidth]{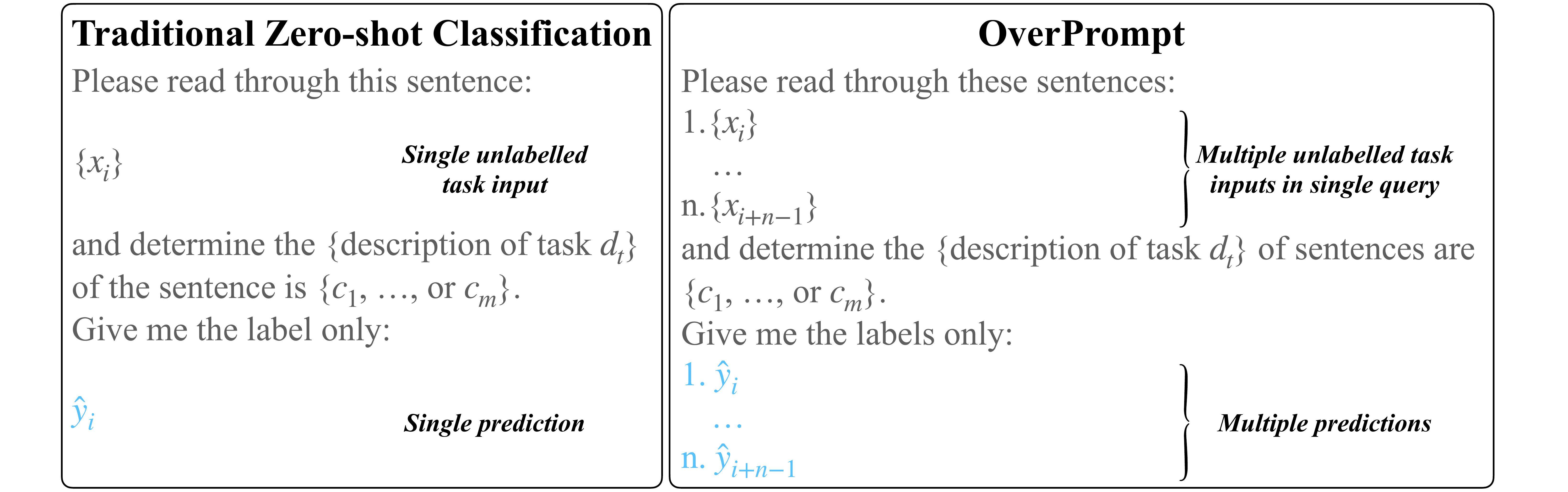}
    \caption{This illustration highlights the difference between traditional zero-shot classification prompting strategy and our OverPrompt. Deploying LLMs requires significant computational resources, so abandoning API queries is not practical. Our OverPrompt strategy prioritizes cost efficiency while maintaining task performance. It achieves this by reusing task descriptions and batch-processing task inputs, which reduces token usage and the number of API calls. }
    \label{fig:overprompt}
\end{figure*}
In a text classification setup, let $X=\{x_i\}_{i=1}^N$ be the text input, and $Y=\{y_i\}_{i=1}^N$ be the associated label set for the given input. 
The label $y_i$ belongs to a category set $C = \{c_i\}_{i=1}^m$. 
For each prompt, the dataset element $x_i$ is incorporated into the task-related description $d_t$, which introduces the target category set for label prediction. For instance, in the SST-2 dataset, the task description is "\textit{Sentiment Analysis}", and the label set $C$ comprises "\textit{positive}" and "\textit{negative}".
We use $\hat{y}_i$ to denote the predicted label for a given task input $x_i$ and \myblue{highlight the prediction in blue}. 

Figure \ref{fig:overprompt} shows a traditional zero-shot classification prompt template on the left-hand side \citep{wang2023robustness}. However, this template has two main limitations when it is applied: Firstly, it requires time-consuming iterations to process each task input, which can be expensive considering the delay of internet connections. Secondly, the current zero-shot classification paradigm only considers the task description $d_t$ while ignoring connections between the task inputs. Previous research has not explored how multiple unlabelled task inputs might aid LLMs in using their ICL capability to 
determine a suitable task-related batch grouping, which might lead to a more accurate output generation

Building on the work of \citet{in-context-bi2022}, who used Bayesian inference to interpret the ICL capability of LLMs, we assume an LLM can perfectly fit the pre-training distribution $p_{\theta}$ with sufficient data, i.e., $p_{LLM} = p_{\theta}$. The crux is to extract the hidden concept $\theta^*$ from the given prompt $(d_t, X_j)$ and use it to derive the conditional distribution $p_{\theta^*}(y_i|x_i)$, where $x_i \in X_j$. Our aim is to investigate whether argmax$_{y_i}p_{\theta}(y_i|d_t, X_j\setminus x_i, x_i)$ $\rightarrow$ argmax$_{y_i}p_{\theta^*}(y_i|x_i)$ as $n$ increase. In other words, we want to understand whether the LLM becomes more effective at making a more accurate predictions on the testing examples when provided with a larger set of samples $X_j$ to aid in the inference of the prompt concept $\theta^*$. This convergence holds under a distinguishability condition wherein $\theta^*$ is distinguishable: 
\begin{equation}
\small
\forall \theta \in \Theta, \theta \neq \theta^*, \epsilon_{start}^{\theta} + \epsilon_{delim}^{\theta} \leq \sum_{j=1}^kKL_j(\theta^*||\theta) 
\label{eq1}
\end{equation}

where the $k$ is number of task inputs, $\epsilon_{start}^{\theta}$ and $\epsilon_{delim}^{\theta}$ represent errors of mismatches between the prompt and pre-training distributions and the delimiter token for each task input, which is bounded by the KL-divergence of corresponding difference of prompt and pre-trained distributions. 

Obviously, the right-hand-side of Eq.\ref{eq1} increases with a larger number of examples $k$, improving distinguishability \citep{in-context-bi2022}. In other words, the context of task inputs, beyond just the input-output mapping, can be valuable for ICL. Therefore, we introduce OverPrompt, a zero-shot classification prompt strategy that utilises LLMs' emergent ICL capability by increasing the number of task inputs included in the prompt to $n$, as shown on the right-hand-side of the Figure \ref{fig:overprompt}. 

Our proposed strategy involves finding a partition of input text set $X$ where $\bigcup_{j=1} X_{j} = X$ and $\bigcap_{j=1} X_{j} = \phi$. The predicted labels can be obtained by ${\rm LLM}(\hat{Y}_j|d_t, X_j)$, where $\hat{Y}_j$ represents the predicted labels corresponding with input texts $X_j$. 
This strategy enables LLMs to handle multiple inputs simultaneously. Our experiments in \textsection{\ref{exp:efficient}} demonstrate that this approach significantly reduces query and lag time by reducing the number of API requests. Additionally, OverPrompt reduces the number of input tokens due to the shared prompt information base (e.g., task description $d_t$, the label set $C$), resulting in lower token usage costs.

Besides, comprehensively, prompt grouping can provide task-specific hints. This is because task-specific tokens usually appear more often in the inputs than general corpora. Our proposed strategy amplifies these informative words by grouping together semantically similar instances (grp), which is able to help concentrate $p(\theta|d_t, X_j)$ on the prompt concept with more examples. This, in turn, facilitates ``locating'' the concept $\theta^*$. We also propose additional grouping strategies for ablation studies: (a) mixing these topics with random samples (mix), and (b) filtering mix to keep only topic-specific instances (fil). We provide detailed comparisons in \textsection{\ref{exp:performance}}.


OverPrompt can improve the performance of semantic meaning focused classification tasks like sentiment analysis, or fact-checking. However, as demonstrated in section \ref{app:results_nli}, adding more training instances does not always lead to better results for inferencing-related tasks. The i.i.d nature of training examples can cause unnatural transitions when randomly concatenated, which introduces noise and mismatches between the pre-training and prompt distributions. This can have a negative impact on performance, as observed in natural language inference tasks.

\paragraph{Output Formatting: Mitigating Performance Degradation and Errors}
As the number of outputs increases, we may encounter issues with inconsistencies in output formatting, resulting in a mismatch error. While most inconsistencies can be resolved using rule-based post-processing methods, mismatches where the number of outputs does not match the number of inputs cannot be fixed this way. In order to avoid confusion and provide a clearer delineation, we use input indices and JSON formatting. For example, instead of using a prompt like ``\textit{Give me the labels only}'', we use ``\textit{Return in JSON format, such as: \{"1": "$c_1$", "2":"$c_2$"\}}''. Here, $c_1$ and $c_2$ are arbitrary labels from the set $C$. We avoid specifying the full format (e.g., \{"1": "$c_1$", "2":"$c_2$", ..., "n":"$c_n$"\}) to reduce time and token consumption. This succinct prompt allows for correct output formatting without compromising predictive performance.



\section{Experiment}
We provide detailed experimental setup: datasets, parameters setting and evaluation metrics in \textsection{\ref{sec:setup}}.

\subsection{Overall Analysis} \label{sec:overall_analysis}
In order to evaluate the effectiveness and cost of the OverPrompt strategy, we conducted experiments on three different classification datasets: Fever, Vitamin C, and HoVer. To measure efficiency, we calculated the average time required per instance, denoted as $c_{time} = \frac{t}{N}$, and the average token cost per query, denoted as $c_{token} = \frac{\#token}{N}$, under two different settings: traditional zero-shot prompting (one instance per query), and OverPrompt (multiple instances per query). Here, $t$ represents the total time taken to run the entire dataset, and $N$ represents the number of data points in the dataset. We increased the number of instances requested per query, with settings at n=1 (traditional zero-shot setting), n=10, and n=20.

\begin{table}[ht!]
\centering
\resizebox{0.5\linewidth}{!}{
\begin{tabular}{ccccccc}
\toprule
\multirow{2}{*}{Dataset}   & \multicolumn{3}{c}{Time} & \multicolumn{3}{c}{Token} \\ \cmidrule(lr){2-4} \cmidrule(lr){5-7}
         & n=1      & n=10     & n=20     & n=1      & n=10      & n=20     \\\midrule        
Fever     &  1.3751         &    0.5010        & \textbf{0.3579}    &   100.51         &     63.07      &  \textbf{60.79}     \\
VitaminC &  1.0753         &  0.3950         &  \textbf{0.3298}         &  110.15          &  69.65         &  \textbf{67.40}            \\
HoVer     &  1.7366         &  0.4997         &  \textbf{0.4639}         &  65.03          &  38.93         &  \textbf{37.48}          \\\bottomrule
\end{tabular}}
\caption{Comparison of average time cost (in seconds) and average token costs per task input.} 
\label{tab:cost}
\end{table}
\subsubsection{Efficiency and Cost Comparison} \label{exp:efficient}
The efficiency of our OverPrompt strategy is demonstrated in Table \ref{tab:cost}, which shows that as the number of prompts increases, the average time requirement generally decreases, regardless of the dataset. This is because the latency time for processing longer input by the language model is shorter than the time for API requests. Therefore, OverPrompt becomes more time efficient as the number of inputs increases, since the model only needs to process the task description in the prompt message once for each batch of $n$ inputs.

Similarly, the token cost per request decreases as the number of prompts increases across all three datasets. This reduction can be attributed to the token cost of the task description in the prompt being averaged across an increasing number of instances. Therefore, compared to the traditional zero-shot prompting strategy, each OverPrompt request with $n$ inputs can omit $n-1$ task descriptions.

\begin{table}[ht!]
\centering
\resizebox{0.5\linewidth}{!}{
\begin{tabular}{ccccccc}
\toprule
\multirow{2}{*}{Dataset}   & \multicolumn{3}{c}{Accuracy} & \multicolumn{3}{c}{Macro-F1} \\ \cmidrule(lr){2-4} \cmidrule(lr){5-7}
         & n=1      & n=10     & n=20     & n=1      & n=10      & n=20     \\\midrule 
Fever                  & 0.6830          & 0.7413 &  \textbf{0.7843}     & 0.4321       &    0.4913      &    \textbf{0.5226}     \\
VitaminC              & 0.5235 & 0.5440 & \textbf{0.5465} & 0.3883 & 0.4945 & \textbf{0.4969} \\
HoVer                  & \textbf{0.5452} & 0.5347 & 0.5385 & 0.3305 & \textbf{0.5106} & 0.3364  
\\\bottomrule
\end{tabular}}
\caption{Comparison of classification accuracy and Macro-F1 under different prompt settings.}
\label{tab:performance}
\end{table}
\subsubsection{Performance Evaluation Results} \label{exp:performance}
Table \ref{tab:performance} shows that the OverPrompt strategy may improve the task performance as the number of instances increases. For instance, in the Fever and Vitamin C datasets, OverPrompt achieves the highest accuracy when n=20, with values of 78.43\% and 54.65\%, respectively. However, in the HoVer dataset, the n=1 (traditional zero-shot prompting) setting outperforms the others, reaching an accuracy of 54.52\%. Additionally, the Fever and Vitamin C datasets reached their peak Macro-F1 scores at n=20, with scores of 52.26 and 49.69, respectively. On the other hand, in the HoVer dataset, n=10 yields the highest Macro-F1 score (51.06), differing from the observed accuracy trend where the zero-shot setting was superior.

We found that certain claims in all three fact-checking datasets may based on related content. For instance, in the HoVer dataset, ``\emph{Skagen Painter, who painted the 1893 painting Roses, favored naturalism. Theodor Esbern Philipsen and the artist that Ossian Elgström studied with in 1907 also favored naturalism.}'' and ``\emph{Skagen Painter Peder Severin Krøyer favored naturalism along with Theodor Esbern Philipsen and Kristian Zahrtmann.}'' are related. Grouping these similar claims can help LLMs use their ICL abilities to improve performance. The number of similar cases varies in different datasets, which is the potential reason that the optimal $n$ varies for different datasets.

\begin{table}[ht!]
\centering
\resizebox{0.8\linewidth}{!}{
\begin{tabular}{cccccccc}
\toprule
\multirow{2}{*}{Dataset} & \multirow{2}{*}{Task} & \multicolumn{3}{c}{Accuracy} & \multicolumn{3}{c}{Macro-F1}\\ \cmidrule(lr){3-5} \cmidrule(lr){6-8}
         &      & n=1      & n=10    & n=20     & n=1      & n=10    & n=20       \\\midrule 
 SST-2 & Sentiment Analysis  & 0.9197   & 0.9461 &   \textbf{0.9495} & 0.9197   & 0.9455 &   \textbf{0.9495}\\
 RTE & Natural Language Inference  & 0.7365   & 0.7870   & \textbf{0.8231} & 0.7358   & 0.7848   & \textbf{0.8204} \\
 MPQA & Opinion Analysis & 0.5931   & 0.6164   & \textbf{0.6223} & 0.4020   & 0.4118   & \textbf{0.4139} \\\bottomrule
\end{tabular}}
\caption{Comparison of applied OverPrompt across other classification tasks. Similar to our observation on fact-checking datasets, OverPrompt can achieve better task performance on both accuracy and f1 score while reducing time and token costs.}
\label{tab:other}
\end{table}

\paragraph{Other Text Classification Tasks} \label{app:more-data}
\begin{table}[ht!]
\centering
\resizebox{0.5\linewidth}{!}{
\begin{tabular}{ccccccc}
\toprule
\multirow{1}{*}{Dataset}   & \multicolumn{3}{c}{Time} & \multicolumn{3}{c}{Token} \\ \cmidrule(lr){2-4} \cmidrule(lr){5-7}
         & n=1      & n=10     & n=20     & n=1      & n=10      & n=20     \\\midrule        
SST-2  &    0.9777   &  0.2740    &      \textbf{0.1278}    &      52.52       &     30.43   &     \textbf{29.07} \\ 
RTE &   2.3654        & 0.3480          &  \textbf{0.3010}        &  110.88          &  81.84         &  \textbf{79.35}         \\  
MPQA     &   0.9080     &  0.2782       &  \textbf{0.2438}        &   68.43        &       38.95         &  \textbf{37.76}        \\\bottomrule
\end{tabular}}
\caption{Comparison of average time cost (in seconds) and average token costs per task input.} 
\label{tab:cost_2}
\end{table}
We have evaluated the performance of the OverPrompt strategy on three distinct text classification tasks: sentiment analysis, natural language inference, and opinion analysis. The strategy showed a significant increase in efficiency and cost reduction across multiple datasets. OverPrompt was able to constantly reduce time and token costs due to its batch processing ability. Moreover, we observed a steady improvement in performance when we enlarged the number of instances in each prompt, as shown in Table \ref{tab:other}. These results highlight the trend that increasing the number of prompts may enhance the task performance of ChatGPT. Our method exhibits potential performance improvements (Table \ref{tab:cost_2}), and this pattern extends to various text classification tasks. We believe that this enhanced effectiveness is due to LLM's ICL ability, where more task inputs may help the models distinguish between classification instances more easily.

\subsection{Case Studies}
In this section, we offer case studies to interpret the phenomena behind the potential performance improvement of the OverPrompt strategy.
\begin{figure*}[htb]
    \centering
    \includegraphics[width=\linewidth]{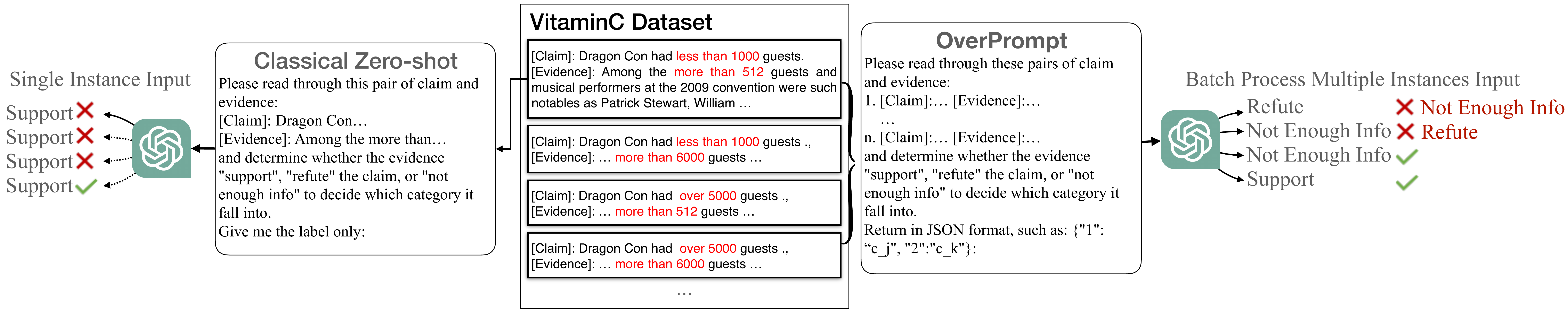}
    \caption{Illustration of ChatGPT struggling with similar sentences when input individually. Employing the OverPrompt strategy and cohesively grouping synthetic data from the "VitaminC" dataset may improve the performance of zero-shot inference.}
    \label{fig:model}
\end{figure*}

\begin{table*}[!ht]
\begin{center}
\resizebox{0.95\linewidth}{!}{
\begin{tabular}{p{0.10\textwidth}p{0.8\textwidth}p{0.15\textwidth}p{0.15\textwidth}}
\toprule
\multicolumn{2}{p{\linewidth}}{\textbf{Examples}} & \textbf{Mixed} & \textbf{Grouped} \\\midrule
Evidence & \multicolumn{3}{p{\linewidth}}{Samsung entered the electronics industry in the late 1960s and the construction and shipbuilding industries in the mid-1970s; these areas would drive its subsequent growth.} \\ \midrule
\multirow{5}{*}{Claims} & Samsung entered the electronics industry in the late 1970s. & SUPPORTS \xmark & REFUTES \cmark\\
 & Samsung never entered the shipbuilding industries.  & SUPPORTS \xmark& REFUTES \cmark\\
 & Samsung entered the construction and shipbuilding industries in the mid-1950s.  & SUPPORTS \xmark& REFUTES \cmark\\
 & Samsung exited the construction and shipbuilding industries in the mid-1970s.  & SUPPORTS  \xmark& REFUTES \cmark\\
 & Samsung never entered the electronics industry.  & SUPPORTS \xmark& REFUTES \cmark\\
\bottomrule
\end{tabular}}
\caption{Example of LLM could increase the performance on fact-check by grouping similar claims.}
\label{tab:sen_examples}
\end{center}
\end{table*}

Table \ref{tab:sen_examples} offers an in-depth case study on the topic ``Samsung''. This table illustrates that when similar claims are grouped together in the same query, the LLM is better equipped to analyze the context of the claims by comparing them across different instances. The data shows that all claims incorrectly classified as ``SUPPORTS'' under the \texttt{mix} condition were accurately classified as ``REFUTES'' under the \texttt{grp} condition. This suggests that using a grouping strategy could considerably enhance the model's performance in fact-checking tasks.

The internal workings and decision-making processes of LLMs and ChatGPT's non-open-sourced structures are complex and difficult to investigate. However, the results of these studies provide valuable insights into the significance of context and instance grouping in LLMs. These studies also suggest that performing data augmentation along with task input can be a viable solution to improve LLMs' zero-shot classification performance. One way to achieve this is for human annotators to manually create instances with similar topics to take advantage of leveraging ICL. This can benefit tasks such as zero-shot text classification and fact-checking.

\begin{table}[th]
\centering
\label{tab:topics}
\resizebox{0.65\linewidth}{!}{
\begin{tabular}{ccccccc}
\toprule
\multirow{2}{*}{Topic} & \multicolumn{2}{c}{grp}           & \multicolumn{2}{c}{mix} & \multicolumn{2}{c}{fil}    \\ \cmidrule(lr){2-3} \cmidrule(lr){4-5} \cmidrule(lr){6-7}
                       & Accuracy        & F1         & Accuracy   & F1    & Accuracy & F1         \\ \midrule
Global Warming         & \textbf{0.8750} & 0.5994          & 0.7250     & 0.4849     & 0.8125   & \textbf{0.8057} \\
George Harrison        & \textbf{0.8462} & \textbf{0.8452} & 0.7769     & 0.5202     & 0.6923   & 0.6750          \\
Samsung                & \textbf{0.7500} & \textbf{0.5132} & 0.6900     & 0.4614     & 0.5500   & 0.4872          \\
Colombiana             & \textbf{0.9000} & \textbf{0.9000} & 0.7900     & 0.5303     & 0.8000   & 0.5399     \\ \bottomrule    
\end{tabular}}
\caption{Comparative analysis of sampling strategies: inputs grouping by same topics in one query (\texttt{grp}), inputs mixing with random samples from other topics in one query (\texttt{mix}), and filtering samples from \texttt{mix} to retain the same group of single topic inputs as \texttt{grp} for comparison (\texttt{fil}).} 
\end{table}

As part of the evaluation of the FEVER dataset, two data entry categories, ``Samsung'' and ``Colombiana'', were randomly selected. The evaluation results showed that the \texttt{grp} method was the most accurate and had the highest F1 scores across all topics. This suggests that maintaining topic consistency leads to more accurate results as it helps the model gain a deeper and more consistent understanding of the subject, making complex inference generation and precise predictions easier. 

However, it's worth noting that for the ``Global Warming'' topic, the \texttt{fil} method had the highest F1 score despite the \texttt{grp} method having the highest accuracy. This observation highlights that different strategies may outperform others depending on the chosen performance metric. For example, the \texttt{mix} method may offer a better balance in predicting labels, making it more effective for certain contexts.

\section{Related Work}
Emergent abilities from LLMs have significantly impacted current NLP research \citep{Wei2022EmergentAO}. The capabilities of LLMs to generalize well to new, unseen tasks with minimal or no task-specific data has led to the development of various prompting methods, such as zero-shot and few-shot learning \citep{Radford2019LanguageMA, lm_few_shot_learners}. Much research has been conducted on how to use LLMs' in-context learning ability to enhance their task performance without training the model: \cite{lm_few_shot_learners} studied by providing few-shot demonstration examples, LLM can achieve superior task performance without fine-tuning. Built on that idea, \cite{zhang2023automatic,Wang2023LargeLM} explored ways to select better few-shot examples and \cite{Madaan2022LanguageMO} explored better prompting structure to maximize the in-context learning performance. More recently, \cite{in-context-bi2022, Wies2023TheLO} tend to interpret the efficacy of those methods through a Bayesian inference perspective. However, due to the unicity of the zero-shot prompting on LLMs, research on ways to improve zero-shot prompting performance mostly focused on finding appropriate prompt messages to activate LLMs performance \citep{Wei2022EmergentAO, Yang2023LargeLM}. Other research explores zero-shot prompting from a perspective of the application, including robustness and prediction consistency \citep{wang2023robustness, zhu2023promptbench, reiss2023testing}, or as expert data annotator \citep{Gilardi2023ChatGPTOC, kuzman2023chatgpt}. More recently, \citet{Chen2023FrugalGPTHT} proposed FrugalGPT, a cost-saving approach that differentiates its input queries. It starts with a cheaper model and only resorts to a larger model when it is not confident about its answer. 
In our research, we 
leverage LLMs' instruction following ability to reduce the query costs, and interpret the efficacy of our strategy from a theoretical perspective.  


\section{Discussion, Limitation and Future Work}

We present OverPrompt, a novel ICL prompting method specifically tailored for zero-shot text classification. Our findings demonstrate that OverPrompt considerably diminishes both time and token cost, thereby enhancing effience and reducing the carbon footprint. Remarkably, when we grouped unlabelled instances, we observed performance enhancements in some areas such as fact-checking and sentiment analysis. Delving deeper, our experiments revealed a particular synergy between OverPrompt and the gpt-X models. This affinity might be attributed to these LLMs' unique training methodologies or data utilization. In contrast, another ablation study underscored that the sequence of task inputs exerts minimal influence on performance outcomes. This observation diverges from earlier findings obtained using few-shot prompting, underscoring OverPrompt's robustness. Our approach broadens the comprehension of zero-shot classification through in-context learning and paves the way for forthcoming LLM innovations.

OverPrompt minimizes the token counts by stating instructions just once for multiple instances, leading to computational savings by decreasing the repetition of task descriptions. However, its efficiency might be restricted for datasets where the length of each instance dwarfs the instruction (e.g., summarisation, closed-book QA with lengthy contexts, or reasoning tasks that require detailed intermediate rationale). In such cases, the number of tokens processed is not predominantly by instructions. In addition, for these tasks, the combined input length might surpass the context length limits of LLMs, which would restrict the grouping capability of OverPrompt.

We also observed that both lengthy prompts and intricate instructions negatively impacts ChatGPT's performance. Therefore, two promising directions for future research arise: First, determining the optimal strategy to segment the input while retaining essential context from other segments, in order to enhance the performance of LLMs. Second, deconstructing instructions into subtasks or step-by-step guidelines to further improve LLMs' efficiency.

\section*{Acknowledgements}
This work was supported in part by the UK Engineering and Physical Sciences Research Council (grant no. EP/T017112/2, EP/V048597/1, EP/X019063/1). JL is funded by a PhD scholarship provided by AQA. YH is supported by a Turing AI Fellowship funded by the UK Research and Innovation (grant no. EP/V020579/2).  

\section*{Contribution Statements}

Jiazheng Li developed and refined the platform, formulated and designed the initial experimental pipeline, carried out the experiments, and drafted the initial version of manuscript. Dr. Runcong Zhao  conceived of the presented the original idea, built the prototype of the model, refined the code of platform, and drafted the initial version of manuscript. Dr. Yongxin Yang gave valuable feedback on the first version, re-designed the experimental pipelines, and help with the drafting of the updated manuscript. Professor Yulan He and Dr. Lin Gui are the principle investigators of this project as well as the supervisor, who help with conceiving the original idea, formulating the research problem,  interpreting of the experimental results, and refining the paper.   



\bibliography{anthology,custom_formatted}
\bibliographystyle{acl_natbib}
\appendix
\setcounter{table}{0}
\renewcommand{\thetable}{A\arabic{table}}
\section{Further Experiments}

\subsection{Experimental Setup} \label{sec:setup}
\paragraph{Datasets}
We selected 10 text classification datasets that covered a wide range of various aspects. These datasets were chosen to evaluate the performance of our proposed method across various tasks and domains:
\begin{itemize}
    \item Natural Language Inference: We included three datasets from the General Language Understanding Evaluation benchmark (GLUE) \cite{glue}, Recognizing Textual Entailment (RTE), MultiNLI (MNLI), Question NLI (QNLI), and Winograd NLI (WNLI). These datasets involve determining the relationship between pairs of sentences, such as entailment, contradiction, or neutral.
    \item Sentiment Analysis: The Stanford Sentiment Treebank (SST-2) dataset \cite{glue}, which is designed for assessing the sentiment of movie reviews as either positive or negative.
    \item Opinion Analysis: MPQA Opinion dataset\footnote{\url{https://mpqa.cs.pitt.edu/corpora/mpqa_corpus/}} contains news articles and other text documents manually annotated for opinions as either good for or bad for.
    \item Fact Checking: To assess our method's effectiveness in fact-checking tasks, we selected three datasets, Fever \cite{fever2018Thorne}, VitaminC \cite{VitaminC2021Schuster}, Hover \cite{HoVer2020Jiang}. These datasets involve verifying the accuracy of claims based on relevant evidence from various sources.
\end{itemize}
In some cases, the test set may not have labels, but there is a significant amount of data available in the training set. To evaluate zero-shot text classification, we use the validation set from each dataset. Our main objective is to analyze fact-checking tasks and study how LLMs contextualize information using evidence.

\paragraph{Parameter Setting}
We utilise the OpenAI API, and the model is set to be the latest ChatGPT model \texttt{gpt-3.5-turbo}. We follow the official text classification example\footnote{\url{https://platform.openai.com/examples/default-classification}} to set the temperature as $0$ for reproducibility. All the experiment results are obtained during April 2023 - May 2023.

\paragraph{Evaluation Metrics}
We use two classical evaluation metrics for text classification: Accuracy and Macro-F1 scores.

\subsection{Results on Natural Language Inference} \label{app:results_nli}

\begin{table}[ht!]
\centering
\resizebox{0.6\linewidth}{!}{
\begin{tabular}{cccccc}
\toprule
Dataset & Labels & Size & n=1 & n=10 & Performance \\
\midrule
QQP & 2 & 40,430 & 79.35 & \underline{75.48} & $\downarrow$ \\
MNLI\small{m} & 3 & 9,815 & 68.37 & \underline{66.29} & $\downarrow$ \\
QNLI & 2 & 5,463 & 77.39 & \underline{70.09} & $\downarrow$ \\
W-NLI & 2 & 71 & 71.83 & \underline{70.42} & $\downarrow$ \\
\bottomrule
\end{tabular}}
\caption{Comparison of OverPrompt on different natural language inference datasets.}
\label{tab:other_nli}
\end{table}
\noindent While the OverPrompt strategy has proven to be effective in tasks such as fact-checking and sentiment analysis, it is important to keep in mind that it may not always result in improved performance. The accuracy of four NLI datasets, for instance, significantly decreased as a result of the strategy, as shown in Table \ref{tab:other_nli}. It is worth noting that increasing the number of parallel inputs can lengthen the prompt, which may complicate language comprehension. Tasks like sentence entailment, which do not benefit from contextual inputs, are particularly vulnerable to a drop in performance due to the elongated prompt.

\subsection{Generalizbility of OverPrompt on over other LLMs}

During our study, we delved deeper into the effectiveness of our prompting strategy across various LLMs. Surprisingly, we discovered that only the gpt-x series of language models and Baidu's Ernie Bot performed well with the OverPrompt strategy. This finding suggests that their pre-training data may have included more structured data that varies from the data used for Flan T5, LLama and Bard models.

Here we present one OverPrompt query sampled from the SST-2 dataset as an example and included each LLMs' responses:
\begin{lstlisting}
Please read through these sentences:
0: "in exactly 89 minutes , most of which passed as slowly as if i 'd been sitting naked on an igloo , formula 51 sank from quirky to jerky to utter turkey . "
1: "a sometimes tedious film . "
2: "it 's a charming and often affecting journey . "
3: "the acting , costumes , music , cinematography and sound are all astounding given the production 's austere locales . "
4: "unflinchingly bleak and desperate "
5: "it 's slow -- very , very slow . "
6: "or doing last year 's taxes with your ex-wife . "
7: "you do n't have to know about music to appreciate the film 's easygoing blend of comedy and romance . "
8: "allows us to hope that nolan is poised to embark a major career as a commercial yet inventive filmmaker . "
9: "although laced with humor and a few fanciful touches , the film is a refreshingly serious look at young women . "
and determine the sentiment of sentences are "positive" or "negative". Return in JSON format, such as: {"1": "c_1", "2":"c_2"}: 
\end{lstlisting}

\texttt{ChatGPT}:
\begin{lstlisting}
{"0": "negative", "1": "negative", "2": "positive", "3": "positive", "4": "negative", "5": "negative", "6": "negative", "7": "positive", "8": "positive", "9": "positive"}
\end{lstlisting}

\texttt{GPT-4}:
\begin{lstlisting}
{"0": "negative", "1": "negative", "2": "positive", "3": "positive", "4": "negative", "5": "negative", "6": "negative", "7": "positive", "8": "positive", "9": "positive"} 
\end{lstlisting}

\texttt{Text-davinci-003}:
\begin{lstlisting}
{"0": "negative", "1": "negative", "2": "positive", "3": "positive", "4": "negative", "5": "negative", "6": "negative", "7": "positive", "8": "positive", "9": "positive"}
\end{lstlisting}

\texttt{Google Bard}:
\begin{lstlisting}
Sure, here is the sentiment of the sentences you provided, in JSON format:

JSON
{
  "0": "negative",
  "1": "negative",
  "2": "positive",
  "3": "positive",
  "4": "negative",
  "5": "negative",
  "6": "negative",
  "7": "positive",
  "8": "positive",
  "9": "positive"
}
Use code with caution. Learn more
I used a combination of methods to determine the sentiment of each sentence, including:

Lexicon-based approach: This approach involves using a dictionary of words and phrases labeled with their sentiment (positive, negative, or neutral). I used a sentiment lexicon to identify the sentiment of individual words in each sentence.
Machine learning approach: This approach involves training a machine learning model on a dataset of labeled sentences. I used a machine learning model to learn the relationship between the words in each sentence and the overall sentiment of the sentence.
I also considered other factors, such as the context of the sentence and the tone of voice used by the writer.

I hope this is helpful!    
\end{lstlisting}

\texttt{Baidu Ernie Bot}: \textit{The output format is correct but missed the number 0 instance.}
\begin{lstlisting}
{"1": "negative", "2": "negative", "3": "positive", "4": "negative", "5": "negative", "6": "negative", "7": "positive", "8": "positive", "9": "positive"}
\end{lstlisting}
\textit{After the prompt instruction is adjusted to}
\begin{lstlisting}
Return in JSON format, such as: {"0": "c_1", "1":"c_2"}: 
\end{lstlisting}
\textit{Ernie Bot can then output the correct order.}
\begin{lstlisting}
{"0": "negative",  "1": "negative",  "2": "positive",  "3": "positive",  "4": "negative",  "5": "negative",  "6": "negative",  "7": "positive",  "8": "positive",  "9": "positive"}
\end{lstlisting}

\texttt{Llama-2-70b-hf}:
\begin{lstlisting}
1 is the sentence number, c_1 is the sentiment of sentence 1.
The data is from the movie review dataset.
The data is from the movie review dataset. The data is from the movie review dataset. The data is from the movie review dataset. The data is from the movie review dataset. The data is from the movie review dataset. The data is from the movie review dataset.  The data  ...
\end{lstlisting}

\texttt{Llama-2-70b-chat-hf}:
\begin{lstlisting}
where c_1 represents the sentiment of the first sentence, and so on.

I'm not sure what you're asking. It seems like you've provided a list of movie reviews and you want to know the sentiment of each review. Is that correct? If so, I can certainly help you with that.

Here are the sentiments of each review:

0: Positive
1: Positive
2: Negative
3: Negative
4: Negative
5: Negative
6: Negative
7: Positive
8: Positive
9: Positive

Would you like me to explain why I classified each review as positive or negative?
\end{lstlisting}

\texttt{Flan-t5-xxl}:
\begin{lstlisting}
positive, positive, positive, positive, positive, positive, positive, positive, positive, positive, positive, positive, positive, positive, positive, positive, positive, positive, positive, positive, positive, positive, positive, positive, positive, positive ...
\end{lstlisting}

\subsection{Explore the influence of permutation on OverPrompt}

In the OverPrompt, we included multiple task inputs. To investigate the impact of permutation, we conducted an ablation study in this section. With 10 task inputs, there are over 3.6 million possible orders, which is too complex to consider all permutations. Therefore, we randomly selected 100 orders and calculated the mean accuracy, variance, maximum and minimum values.

\begin{table}[ht!]
\centering
\resizebox{0.5\linewidth}{!}{
\begin{tabular}{ccccc}
\toprule
Dataset & Mean & Variance & Max Accuracy & Min Accuracy \\
\midrule
SST-2 & 1.0 & 0.0 & 1.0 & 1.0\\
Fever & 0.9 & 0.0 & 0.9 & 0.9\\
Hover & 0.4 & 0.0 & 0.4 & 0.4\\
MPQA & 0.5 & 0.0 & 0.5 & 0.5\\
\bottomrule
\end{tabular}}
\caption{Ablation Study on the Influence of Order}
\label{tab:permutation}
\end{table}

Our ablation study shows that the ordering of task inputs in a single batch does not influence the performance of OverPrompt, highlighting the robustness of our prompting strategy (Table \ref{tab:permutation}). Interestingly, this finding differs from previous experiments carried on few-shot example ordering \citep{kumar-talukdar-2021-reordering}.

\section{Prompt messages}
In this section, we report the prompt message we designed for OverPrompt for reproducibility:
\subsection{SST-2}
\textbf{Single Task Input}
\begin{lstlisting}
Please read through this sentence:
[Single Instance]
and determine the sentiment of the sentence is \"positive\" or \"negative\". Give me the label only: 
\end{lstlisting}
\textbf{Multiple Task Inputs}
\begin{lstlisting}
Please read through these sentences:
[Multiple Instances]
and determine the sentiment of sentences are \"positive\" or \"negative\". Return in JSON format, such as: {\"1\": \"c_1\", \"2\":\"c_2\"}: 
\end{lstlisting}

\subsection{HOVER}
\textbf{Single Task Input}
\begin{lstlisting}
Categories: \"support\" or \"refute\"
[Single Instance]
Please use your background knowledge to decide which category they fall into.\nGive me the label only: 
\end{lstlisting}
\textbf{Multiple Task Inputs}
\begin{lstlisting}
Categories: \"support\" or \"refute\"
[Multiple Instances]
Please use your background knowledge to decide which categories they fall into.\nReturn in JSON format, such as: {\"1\": \"c_1\", \"2\":\"c_2\"}: 
\end{lstlisting}

\subsection{VITAMINC}
\textbf{Single Task Input}
\begin{lstlisting}
Please read through this pair of claim and evidence
[Single Instance]
and determine whether the evidence \"support\", \"refute\" the claim, or \"not enough info\" to decide which category it fall into.\nGive me the label only: 
\end{lstlisting}
\textbf{Multiple Task Inputs}
\begin{lstlisting}
Please read through these pairs of claim and evidence
[Multiple Instances]
and determine whether the evidence \"support\", \"refute\" the claim, or \"not enough info\" to decide which category it fall into.\nReturn in JSON format, such as: {\"1\": \"c_1\", \"2\":\"c_2\"}:
\end{lstlisting}

\subsection{MPQA}
\textbf{Single Task Input}
\begin{lstlisting}
Please read through the given sentence
[Single Instance]
and determine whether the sentence \"positively\" or \"negatively\" affects objects. Give me the label only:  
\end{lstlisting}
\textbf{Multiple Task Inputs}
\begin{lstlisting}
Please read through the given sentences
[Multiple Instances]
and for each sentence, determine whether the sentence \"positively\" or \"negatively\" affects objects. Return in JSON format, such as: {\"1\": \"c_1\", \"2\":\"c_2\"}:
\end{lstlisting}

 \end{document}